\pdfoutput=1

\documentclass[10pt, a4paper]{article}
\usepackage{lrec-coling2024} %

\usepackage{booktabs}

\title{
CoNLL\#: Fine-grained Error Analysis and a Corrected Test Set for CoNLL-03 English*\thanks{* This is a preprint version of this article. Please cite \href{https://aclanthology.org/2024.lrec-main.330/}{the version in the LREC-COLING 2024 proceedings}.}
}

\name{Andrew Rueda$^{\ast}$, Elena Álvarez Mellado$^{\dagger}$, Constantine Lignos$^{\ast}$} 

\address{$^{\ast}$Michtom School of Computer Science, Brandeis University\\
    $^{\dagger}$NLP \& IR Group, School of Computer Science, UNED\\
     \texttt{\{andrewrueda,lignos\}@brandeis.edu}\\
     \texttt{elena.alvarez@lsi.uned.es}\\
}

\abstract{
Modern named entity recognition systems have steadily improved performance in the age of larger and more powerful neural models. However, over the past several years, the state-of-the-art has seemingly hit another plateau on the benchmark CoNLL-03 English dataset. In this paper, we perform a deep dive into the test outputs of the highest-performing NER models, conducting a fine-grained evaluation of their performance by introducing new document-level annotations on the test set. We go beyond F1 scores by categorizing errors in order to interpret the true state of the art for NER and guide future work. We review previous attempts at correcting the various flaws of the test set and introduce CoNLL\#, a new corrected version of the test set that addresses its systematic and most prevalent errors,  allowing for low-noise, interpretable error analysis.
\\ \newline  }

\begin{document}

\maketitleabstract

\section{Introduction}
In recent years, we have seen substantial improvements in named entity recognition (NER) performance due to new techniques such as pretrained language models, novel approaches that use document-level context and entity embeddings, and a general increase in computational power.
These advancements have resulted in a steady rise in state-of-the-art performance on the usual NER benchmarks, with the CoNLL-03 English corpus \citeplanguageresource{tjong-kim-sang-de-meulder-2003-introduction} being the most popular evaluation set.
However, results on these NER benchmarks seem to have plateaued since 2021, and a glass ceiling for the task has been hypothesized \citep{stanislawek-etal-2019-named}. This raises the question: what is there left to improve for the CoNLL-03 English NER task?

In this paper, we investigate what NER models are still struggling with. In order to do so, we run three of the best performing models for NER on the CoNLL-03 English dataset and conduct a fine-grained error analysis that goes beyond the traditional false positive/false negative distinction that informs span-level F1 score. 
The goal of this analysis is to hone in on the lingering errors from state-of-the-art NER models.

The first step was to annotate the 231 original documents in the test set by assigning each a document domain and a document format. 
We train and test three models with recent state-of-the-art F1 scores and report their performances across document domains and formats.
Due to the presence of significant annotation errors in the original CoNLL-03 English corpus, we gathered the best known gold label corrections (Section \ref{section:previous_work}), and went through multiple rounds of adjudication to fix data processing errors pervasive in the test set, including faulty sentence boundaries and tokenization errors.

The result is CoNLL\#, a new version of the CoNLL-03 English test set which corrects issues more consistently than previous attempts.\footnote{CoNLL\# is released at \url{https://github.com/bltlab/conll-sharp}.}
We rescore the models using this corrected set, showing improved performance across the board.
Finally, we carry out a quantitative and qualitative error analysis to find interpretable, meaningful patterns among the models' prediction errors.

The contributions of this paper are as follows.
First, we report results on CoNLL-03 English for three state-of-the-art NER models and document their fine-grained performance across different document types (formats and domains).
Second, we release a revised version of the CoNLL-03 English test set (CoNLL\#) that incorporates the adjudicated corrections made in previous revised versions, along with corrections for the systematic errors we identified that none of the previous versions fixed.
Finally, we report results of the three SOTA NER models on the new revised CoNLL\# dataset. Our error analysis (both qualitative and quantitative) sheds light on the issues that SOTA models are still struggling with.

\section{Previous Work}
\label{section:previous_work}

Recent advances in NER, such as using document-level features \citep{schweter2020flert} and entity embeddings \citep{yamada-etal-2020-luke} have improved results on popular NER benchmarks, with the best reported F1 scores on CoNLL-03 English surpassing 94.
However, after decades of steady progress in NER, recent models seem to have reached a plateau. 

Prior work has already pointed out the potential existence of a glass ceiling that looms over NER \citep{stanislawek-etal-2019-named}. 
Errors in the gold standard have been pointed out as a possible cause. 
As a result, there have been significant prior efforts to identify annotation errors and release corrected versions of the data.
\citet{stanislawek-etal-2019-named} investigated errors in state-of-the-art NER models using linguistic knowledge to categorize errors within an original annotation schema for all entity types except MISC. They found that models at the time struggled with predictions that required gathering sentence-level context, as well as document-level co-reference.

CoNLL++ \citeplanguageresource{wang-etal-2019-crossweigh} is at this point the most widely-adopted corrected test set for CoNLL-03 English.
They introduced the CrossWeigh framework to identify potential label mistakes in NER data. CoNLL++ itself corrects 309 labels in the original test set in a way that remains consistent with the original annotation guidelines which date back to the MUC-7 Named Entity task.\footnote{\label{note2}\url{https://www-nlpir.nist.gov/related_projects/muc/proceedings/ne_task.html}}
The authors made the decision not to correct any of the sentence boundaries or tokenization errors present in the test set.

ReCoNLL \citeplanguageresource{Fu_Liu_Zhang_2020} was a similar attempt that used a measure called Entity Coverage Ratio to identify mentions that appeared in the test set with different labels than in the training set, manually correcting those labels when needed.
This resulted in 105 corrected labels and 10 corrected sentence boundaries in the test set.

\citetlanguageresource{reiss-etal-2020-identifying} embarked on a wider-scale correction effort, using a self-supervised approach to alter 1,320 token labels across the entire corpus.
This included effectively making changes to the annotation guidelines to make the annotation more consistent.
These changes, which resulted in a corrected set we will refer to as CoNLL-CODAIT, included a concerted effort to correct faulty sentence boundaries and token errors.

After this paper was submitted for review, another corrected version of CoNLL-03 English was released: \textsc{CleanCoNLL} \cite{rucker-akbik-2023-cleanconll}. 
\textsc{CleanCoNLL} is a similar relabeling effort that seeks to fix the existing errors in CoNLL-03 English.
\textsc{CleanCoNLL} was derived from CoNLL-CODAIT, and consequently, the training set and development set were also modified.
In contrast, our approach tries to conform to the original annotation guidelines and does not modify the training and development sets.
As this work was performed concurrently with ours, we do not discuss \textsc{CleanCoNLL} in detail, but Section \ref{sec:adjudication} provides counts for the number of tokens in disagreement between CoNLL\#, \textsc{CleanCoNLL}, and CoNLL-CODAIT.

Despite these many efforts, corrected versions do not always agree with one another, and the question of what are NER models struggling with still remains. 
In this paper, we conduct a fine-grained analysis to identify systematic errors that SOTA NER models make on the CoNLL-03 English benchmark in order to better understand what may be the reason for this recent plateau.
We also identify systematic bugs with the original CoNLL-03 English test set and propose a new corrected version that adjudicates and ameliorates previous versions.

\section{Document Type Annotation}
\label{sec:document_types}

In order to get a general overview of the type of errors models could be struggling with, we first annotated each of the documents in the CoNLL-03 English test set according to its type.
Each document was annotated with two labels: one for the domain or genre of the news piece, and another for the document format (explained below).
The motivation behind this decision was that CoNLL-03 English suffers from data selection  biases, most notably the overrepresentation of sports articles compared to other domains \citep{chiticariu-etal-2010-domain,nagesh-etal-2012-towards}.
By annotating the domain and format on the document level, we wanted to see if models were making systematic errors on certain document types. 
Table \ref{tab:document_types} summarizes the number of documents in the CoNLL-03 English test set per domain and format.

\subsection{Document Domains}
\label{subsection:document_domains}
Documents in CoNLL-03 English were extracted from online newswire articles in December 1996.
Beyond the well-known skew toward sports articles, \citet{Agerri2016-fy} also note the inclusion of ``lots of financial and sports data in tables,'' but to our knowledge no one has quantified these findings on the document level.
We annotated the 231 test documents and found that the documents fell neatly within 3 major categories:

\paragraph{Sports} 101 out of 231 documents were indeed news items covering international sporting events. 
These documents were a mixture of text news stories, sports schedules, and scores extracted from tabular data (see Section \ref{subsection:document_formats}). 

\paragraph{Economy} 67 of the documents involved international trade, business, and market updates. 
Similarly to sports, some of the articles in this domain were primarily tabular data (stock indexes, market prices, etc.).

\paragraph{World Events} 63 documents involved general news and events from around the world. This includes politics, crime, natural disasters, etc. Though slightly varied in nature, they share the quality that none of them are in the aforementioned data formats, and are all text-based articles written in complete sentences.

\subsection{Document Formats}
\label{subsection:document_formats}
The documents in CoNLL-03 English test set were annotated into the following three categories:

\paragraph{Text article} News articles written in complete English sentences.

\paragraph{Data report} News reports that, beyond having an informative title, consisted solely of strings extracted from tabular data.

\paragraph{Hybrid} Articles that had a section written in text, as well as a section in data format.\\

Below is an example of part of a document we labeled with the sports domain and data report format:\\
\texttt{
Chelsea B-ORG \\
2 O \\
Everton B-ORG\\
2 O \\
Conventry B-ORG \\
1 O \\
Tottenham B-ORG \\
2 O
}

\begin{table}[tb]
\small
\centering
\begin{tabular}{lrrrrr}
\toprule
& \multicolumn{4}{c}{Domain}\\
\cmidrule(lr){2-5}
             & World  & Economy & Sports & Total\\
Format       & Events &  &  & \\
\midrule
Text Article & 63 & 45 & 31 & 139 \\
Data Report & 0 & 14 & 59 & 73 \\
Hybrid & 0 & 8 & 11 & 19 \\
\midrule
Total & 63 & 67 & 101 & 231 \\
\bottomrule
\end{tabular}
\caption{Test set documents per domain and format}
\label{tab:document_types}
\end{table}

\section{Modeling}
\label{sec:modeling}
We then trained and tested three NER SOTA models on the CoNLL-03 English dataset. 
The document-level annotation described on Section \ref{sec:document_types} was only  devised for the error analysis step and performed on the test set, so it was not used in any way during the training of the models.
\subsection{Models}
We chose three of the best NER models that have recently pushed the state-of-the-art results, each with a reported F1 score greater than 93.0.

\paragraph{XLM-R FLERT} Cross-lingual RoBERTa embeddings fine-tuned on CoNLL-03 English using document context \citep{schweter2020flert}. We trained this model using the published best configuration and the provided random seed.

\paragraph{LUKE} Embeddings that represent both words and ``entities'' (contextualized mentions-strings) using the BERT Masked Language Model objective \citep{yamada-etal-2020-luke}. We carried out the steps given by the authors to load their best model.\footnote{\url{https://colab.research.google.com/github/studio-ousia/luke/blob/master/notebooks/huggingface_conll_2003.ipynb}}

\paragraph{ASP-T0-3B} A 3-billion parameter T5 model fine-tuned using a novel structure-building approach to capture dependencies \citeplanguageresource{liu-etal-2022-autoregressive}. We trained this model using 10 random seeds, and extracted the best test labels.\footnote{The original paper provides contradictory information on the best configuration. Our experiments yielded higher performance on ASP-T0-3B than on ASP-T5-3B.}\\

All three models were trained on the \emph{original} (non-corrected) CoNLL-03 English training set.
For each model, we followed the configurations made available by the authors. 
We trained XLM-R FLERT and ASP-T0-3B and when necessary, augmented the original code in order to extract the per-token predicted labels.

\subsection{Results}
\label{subsection:performance}

\begin{table}[tb]
    \small
    \centering
    \begin{tabular}{lrrrr}
        \toprule
        Model & Precision & Recall & F1 \\
        \midrule
        XLM-R FLERT & 92.87 & \textbf{94.53} & 93.64 \\
        LUKE & \textbf{95.64} & 94.51 & \textbf{94.44} \\
        ASP-T0-3B & 93.65 & 94.15 & 93.88 \\
        \bottomrule
    \end{tabular}
    \caption{Replicated results obtained on the original CoNLL-03 English test set}
    \label{tab:results_original}
\end{table}

\begin{table*}[tb]
    \small
    \centering
    \begin{tabular}{lrrrr}
    \toprule
         & Sports & World Events & Economy & All domains \\
         \midrule
        XLM-R FLERT &  &  &  & \\
        \hspace{0.25cm}Text Article & 94.09 & 94.62 & 90.75 & 93.37 \\ 
        \hspace{0.25cm}Data Report & 94.94 & - & 74.48 & 93.64 \\
        \hspace{0.25cm}Hybrid & 97.51 & - & 77.31 & 95.45 \\
        \hspace{0.25cm}All Formats & 95.17 & 94.62 & 87.68 & 93.69 \\ 
        \midrule
        LUKE & &  & &  \\ 
        \hspace{0.25cm}Text Article & 94.61 & 94.72 & 91.02 & 93.62 \\ 
        \hspace{0.25cm}Data Report & 95.52 & - & 75.44 & 94.27 \\ 
        \hspace{0.25cm}Hybrid & 99.33 & - & 97.35 & 99.14 \\ 
       \hspace{0.25cm}All Formats & 95.93 & 94.72 & 89.22 & 94.44 \\ 
       \midrule
        ASP-T0-3B &  &  &  &  \\ 
        \hspace{0.25cm}Text Article & 91.8 & 95.55 & 90.69 & 93.23 \\ 
        \hspace{0.25cm}Data Report & 94.56 & - & 84.98 & 93.98 \\ 
        \hspace{0.25cm}Hybrid & 97.68 & - & 90.63 & 96.83 \\ 
        \hspace{0.25cm}All Formats & 94.48 & 95.55 & 89.93 & 93.88 \\ 
        \bottomrule
    \end{tabular}
    \caption{CoNLL-03 English test set F1 across document formats and domains}
    \label{tab:results_original_types}
\end{table*}

Table \ref{tab:results_original} summarizes the performance of each model on the CoNLL-03 English test set.
Table \ref{tab:results_original_types} displays results split across domains and formats, using the annotation from Section \ref{sec:document_types}.
These results show that, despite attention paid to sports articles in past work, the economy domain is in fact the lowest-performing domain. This will be explored in greater detail in Section \ref{section:error_analysis}.

\section{Towards CoNLL\#: Adjudicating among Previous Corrections and Making Additional Corrections}
\label{sec:adjudication}

A cursory analysis of the output produced by the models from Section \ref{sec:modeling} revealed that some of the errors were not in fact the models' fault, but instead annotation mistakes in the gold standard.
Errors in the annotation prevent us from doing a reliable diagnosis on what NER SOTA models are really struggling with.
Therefore we decided to manually correct these annotation errors on the CoNLL-03 English test set and rerun the models on a cleaner version of the test set, so that our results would truly illuminate the CoNLL-03 English instances that NER models find most challenging. 

We partly based our corrections on  previously-published corrected versions of the CoNLL-03 English test set: CoNLL++ \citeplanguageresource{wang-etal-2019-crossweigh}, ReCoNLL \citeplanguageresource{Fu_Liu_Zhang_2020}, and CoNLL-CODAIT \citeplanguageresource{reiss-etal-2020-identifying}.
We decided to perform an adjudication process among all three corrected test sets, comparing labels across versions, and making final decisions when they disagreed.

These three corrected versions of CoNLL-03 English took very different approaches to the correction process. 
Consequently, their results vary greatly. %
For instance, CoNLL++ and ReCoNLL did not attempt any token corrections, such as repairing token typos, or splitting faulty tokens.
As a result, they have a perfect one-for-one token overlap, in contrast with CoNLL-CODAIT, which sought to fix all token errors, as well as faulty sentence boundaries (see Table \ref{tab:tokens_diff}).

\begin{table*}[tb]
    \small
    \centering
\begin{tabular}{lrrrr}
        \toprule
        ~ & CoNLL++ & ReCoNLL & CoNLL-CODAIT & CoNLL\# \\
        \midrule
        CoNLL-03 & 309 & 105 & 565 & 457 \\ 
        CoNLL++ & ~ & 276 & 544 & 261 \\
        ReCoNLL & ~ & ~ & 599 & 360 \\ 
        CoNLL-CODAIT & ~ & ~ & ~ & 494 \\ 
        \bottomrule
    \end{tabular}
\caption{Differences in token labels across test sets}
\label{tab:tokens_diff}
\end{table*}

\begin{table*}[ht]
    \small
    \centering
    \begin{tabular}{lrl}
    \toprule
        Error fix & Count & Example \\ 
        \midrule
        Token splits   & 5 &  \textit{JosepGuardiola} $\rightarrow$ \textit{Josep Guardiola} \\ 
        Bad hyphen fixes        & 27 & \textit{SKIING-WORLD CUP} $\rightarrow$ \textit{SKIING - WORLD CUP} \\ 
        Sentence boundary fixes & 63 & [\emph{Results of National Basketball}] [\emph{Association games on Friday}] \\ 
        Label fixes   & 457 & \textit{Tasmania} \texttt{LOC} $\rightarrow$ \textit{Tasmania}  \texttt{ORG} \\ 
        \bottomrule
    \end{tabular}
    \caption{Total number of fixes per type in CoNLL\# compared to CoNLL-03}
    \label{tab:fixes_per_type}
\end{table*}

With this in mind, we decided to proceed with our adjudication process as follows.
First, we compared the labels of CoNLL++ and ReCoNLL, making adjudication decisions for each of their disagreements.
Second, we used CoNLL-CODAIT as a starting point for correcting \emph{all} of the sentence and token boundary errors in the test set.
Finally, we compare token-level label disagreements with our new test set and CoNLL-CODAIT.

The result of this process is CoNLL\#, a corrected version of the CoNLL-03 English test set that adjudicates disagreements among previous corrected versions and includes fixes for errors none of the previous versions considered.

\subsection{Comparing CoNLL++ and ReCoNLL}
CoNLL++ and ReCoNLL had 276 token-level disagreements. 
We compared them side by side and manually adjudicated them. 
CoNLL++ was generally more aggressive in its relabeling efforts, but did so with high precision, as 68.95\% of the disagreements were judged to be in favor of CoNLL++, with ReCoNLL having the correct label in 27.44\% of disagreements, and 10 cases (3.61\%)  where neither were correct.

Highlighting some examples, CoNLL++ did a better job of correcting the labels of domestic sports organizations to ORG, such as the \emph{Tasmania} and \emph{Victoria} Australian Rugby clubs, as well as an Egyptian soccer team with the nickname \emph{ARAB CONTRACTORS}. CoNLL++ also had superior labels for MISC mentions, such as properly labeling \emph{Czech} as a MISC in the sentence \emph{Czech ambassador to the United Nations, Karel Kovanda, told the daily media...}

We also identified an invalid label transition in the ReCoNLL dataset (from \texttt{O} to \texttt{I-PER}), by using SeqScore's \citep{palen-michel-etal-2021-seqscore} validation.

\subsection{Repairing Token and Sentence Boundary Errors}

Many of the incorrect labels in the CoNLL-03 English test set stem from sentence boundary errors in the original test set.
For example, the following sentence boundary (shown with brackets) in the CoNLL-03 English test set interrupts the mention \emph{National Basketball Association}, making it impossible to have a single mention that spans both sentences: [\emph{Results of National Basketball}] [\emph{Association games on Friday}].

CoNLL++ did not attempt to correct any of these sentence boundary errors. In the test set, ReCoNLL corrected 10 sentence boundaries, and CoNLL-CODAIT corrected 26.

For CoNLL\#, we attempted to fix all of sentence boundary errors in the 231 test documents---whether or not they happened to interrupt a mention---to allow NER models to be able to predict on correct sentence boundaries.
We used CoNLL-CODAIT's sentence correction as a starting point, but through manual effort, found many more errors.
We found a systematic sentence boundary error among documents that were labeled in our document-level annotation as sports data reports.
Among the 59 test documents that were sports data reports, 43 of them had a faulty sentence boundary in their initial headline between the 16th and 20th characters. 
This processing error is not observed within any other document type. 
For CoNLL\#, we repaired all 70 sentence boundary errors we identified in the test set.

A similar, systematic error was also found within sports data reports.
The sports headlines from these documents feature a hyphen between the name of the sport and the headline of the article, such as the token \emph{SKIING-GOETSCHL} in the test sentence \emph{ALPINE SKIING-GOETCHL WINS WORLD CUP DOWNHILL}.
If this hyphen were treated as intended, effectively as a colon, the token \emph{GOETCHL} should have been labeled as \texttt{B-PER}.

This type of error occurred 27 times in the original test set, almost all within sports data reports.
Only CoNLL-CODAIT attempted to fix these tokenization errors.
Our manual inspection found that CoNLL-CODAIT corrected 14 out of 27 of these tokens; for CoNLL\#, we fixed all 27 errors.

CoNLL-CODAIT also departed from the CoNLL-03 English tokenization and annotation guidelines by splitting some hyphen-joined entities into two.
For example, in the original dataset \emph{UK-US} in a context like \emph{UK-US open skies talks end} should be a single token annotated as \texttt{B-MISC}.
CoNLL-CODAIT changed 8 instances of tokens like this to be three tokens (\emph{UK - US}), annotated as \texttt{B-LOC O B-LOC}.
In CoNLL\#, we maintained the original tokenization and labels for these tokens.

Overall, CoNLL-CODAIT made 68 corrections on the test set with regards to sentence boundaries and tokenization, of which we accepted 60 in our adjudication process.
CoNLL\# contains an additional 42 similar corrections.

\subsection{Comparing Labels to CoNLL-CODAIT}
\label{subsection:CoNLL_CODAIT}
CoNLL-CODAIT included a far-reaching reannotation project for each of the CoNLL-03 English training, development, and test sets.
For example, they changed national sports teams from LOC to ORG in sentences like \emph{Japan began the defense of their Asian Cup Title with a lucky 2-1 win against Syria in a Group C Championship match on Friday.}
Following the original annotation guidelines, local sports teams that are referred to using a location should be annotated as ORG, while national sports teams referred to using a country name should be annotated as LOC.\footnote{See section A.2.2 from the original MUC-7 guidelines: \url{https://www-nlpir.nist.gov/related_projects/muc/proceedings/ne_task.html}.}
The change to annotate both as ORG reduces the number of arbitrary distinctions models must make, but is not in keeping with the original annotation guidelines.

Table \ref{tab:tokens_diff} shows that at a token level, CoNLL-CODAIT labels have many more disagreements relative to the other test sets in question. 
This is due to the fact that the CoNLL-CODAIT approach included not just correcting annotation errors, but also changing the annotation guidelines.

For instance, as noted by both the authors of CoNLL++ and CoNLL-CODAIT, there are 41 examples in which upcoming sports games displayed in the common format of \emph{ANAHEIM AT BUFFALO} labeled the latter team as LOC. 
This labeling decision is present in all of the CoNLL-03 English training, dev, and test data, and so a proper correction should either overturn all of these labels in the corpus (as CoNLL-CODAIT did), or leave all of them the same (as CoNLL++ did).

Although there are differences between what the annotation guidelines required and what the CoNLL-03 English annotators did, our analysis suggests that no single interpretation of schema for NER is infallible, and are in many cases subjective for tough or ambiguous mentions. 
Given that we did not aim to modify any annotations in the training data, we decided to not change the annotation guidelines governing the test data.
This approach ensured that the labeling decisions implemented in the test data would be in line with the annotation decisions in the training data, contributing to low-noise error analysis. 
As a result, our round of label adjudication with CoNLL-CODAIT maintained most of the labels decided upon in our first round of adjudication between CoNLL++ and ReCoNLL: we adopted the CoNLL-CODAIT label in only 4.86\% of the disagreements.

The result of these two rounds of adjudication, repairs of tokens and sentence boundaries, and arbitration of label disagreements among the corrected test sets, is CoNLL\#, our corrected version of the CoNLL-03 English dataset.
Table \ref{tab:fixes_per_type} summarizes the number of fixes per type in CoNLL\# compared to CoNLL-03.
Token-level comparison between CoNLL\# and the other test sets can be seen in Table \ref{tab:tokens_diff}. 

\begin{table}[tb]
    \small
    \centering
    \begin{tabular}{p{2.5cm}p{2.5cm}r}
    \toprule
        Agreed & Disagreed & Count \\ 
        \midrule
        CoNLL-CODAIT \newline CleanCoNLL  \newline CoNLL\# &  & 49,593 \\
        \midrule
                & CoNLL-CODAIT \newline CleanCoNLL  \newline CoNLL\# & 15 \\ 
                \midrule
        CoNLL-CODAIT  \newline CleanCoNLL & CoNLL\# & 291 \\ 
        \midrule
        CleanCoNLL  \newline CoNLL\# & CoNLL-CODAIT & 180	 \\
        \midrule
        CoNLL-CODAIT  \newline CoNLL\# & CleanCoNLL & 316	 \\ 
        \bottomrule
    \end{tabular}
    \caption{Differences between \textsc{CleanCoNLL}, CoNLL-CODAIT, and CoNLL\#}
    \label{tab:cleanconll_diff}
\end{table}

Table \ref{tab:cleanconll_diff} summarizes the types and number of label disagreements between \textsc{CleanCoNLL}, CoNLL-CODAIT and CoNLL\#.

\section{Results and Error Analysis on CoNLL\#}
\label{section:error_analysis}

We reran the three SOTA NER models from Section \ref{sec:modeling} on our new corrected CoNLL-03 English test set, CoNLL\#, and evaluated the results.
Table \ref{tab:results_conllsharp} shows the results obtained on the new CoNLL\# test set compared to previous results obtained on the original CoNLL-03 test set.
Table \ref{tab:recall_unseen} shows recall results over previously seen (during NER training) and unseen entities.
With the new CoNLL\#, the overall F1 for each of the models increased by more than 2 points.

\begin{table}[tb]
    \small
    \centering
    \begin{tabular}{lrr}
    \toprule
        Model & CoNLL-03 & CoNLL\# \\ 
        \midrule
        XLM-R FLERT & 93.64 & 95.98 \\ 
        LUKE        & \textbf{94.44} & \textbf{97.10} \\ 
        ASP-T0-3B   & 93.88 & 96.50 \\ 
        \bottomrule
    \end{tabular}
    \caption{F1 scores for 3 SOTA models on CoNLL-03 and CoNLL\#}
    \label{tab:results_conllsharp}
\end{table}

\begin{table}[tb]
\small
\centering
\begin{tabular}{lrrrr}
\toprule
& \multicolumn{2}{c}{CoNLL-03} & \multicolumn{2}{c}{CoNLL\#}\\
\cmidrule(lr){2-3}
\cmidrule(lr){4-5}
             & Seen  &  Unseen &  Seen  &  Unseen\\
\midrule
XLMR & 96.49 & \textbf{92.36} & 98.47 & 93.98\\
LUKE & \textbf{96.90} & 91.88 & \textbf{99.22} & \textbf{94.64} \\
ASP-T0-3B & 96.42 & 91.64 & 98.57 & 94.63 \\
\bottomrule
\end{tabular}
\caption{Recall results for seen and unseen mentions en CoNLL-03 and CoNLL\#}
\label{tab:recall_unseen}
\end{table}

\begin{table*}[tb]
    \small
    \centering
    \begin{tabular}{lrrrr}
    \toprule
    & Sports & World Events & Economy & All Domains \\
         \midrule
        XLM-R FLERT &  &  &  & \\
        \hspace{0.25cm}Text Article & 95.00 & 97.18 & 93.59 & 95.61 \\ 
        \hspace{0.25cm}Data Report & 97.72 & - & 78.38 & 96.46 \\ 
        \hspace{0.25cm}Hybrid & 98.09 & - & 76.27 & 95.88 \\ 
        \hspace{0.25cm}All Formats & 97.22 & 97.18 & 90.43 & 95.94 \\ 
        \midrule
        LUKE &  &  &  & \\ 
        \hspace{0.25cm}Text Article & 96.70 & 97.54 & 95.26 & 96.68 \\ 
        \hspace{0.25cm}Data Report & 98.42 & - & 76.66 & 97.05 \\ 
        \hspace{0.25cm}Hybrid & 99.81 & - & 95.58 & 99.40 \\ 
        \hspace{0.25cm}All Formats & 98.29 & 97.54 & 92.67 & 97.10 \\
        \midrule
        ASP-T0-3B  &  &  &  & \\
        \hspace{0.25cm}Text Article & 93.34 & 97.74 & 95.32 & 95.97 \\ 
        \hspace{0.25cm}Data Report & 97.96 & - & 81.91 & 96.92 \\ 
        \hspace{0.25cm}Hybrid & 98.22 & - & 90.27 & 97.46 \\ 
        \hspace{0.25cm}All Formats & 97.05 & 97.74 & 93.13 & 96.50 \\ 
    \bottomrule
    \end{tabular}
    \caption{Model performance across domain and format on CoNLL\#}
    \label{tab:results_conllsharp_docs}
\end{table*}

Table \ref{tab:results_conllsharp_docs} summarizes the results obtained across document domains and document formats.
The performance gains made by testing on CoNLL\# seem to have a roughly uniform effect across all of the different document types. 
The predictions on economy documents have improved, but these documents are still the predominant source of lingering errors. 
In fact, while performance on economy text articles increased by multiple points for each model, their respective performances on data reports and hybrid articles went \emph{down}, which is unique to those two document types.

We conducted an annotated error analysis on outputs of the three state-of-the-art English NER models. 
This was done in a side-by-side analysis of the gold token-level BIO labels and the predicted labels in spreadsheet format.  
For each token-level label mismatch, context of the nearby tags was used to classify each error within the following schema from \citet{chinchor-sundheim-1993-muc}:
 \textbf{Missed} for a full false negative,
 \textbf{Spurious} for a full false positive,
 \textbf{Boundary Error} when a mention was detected but with imperfect overlap, and
 \textbf{Type Error} for when the boundaries were perfect but the tag type was incorrect.
 For each Type Error, the sub-type was also recorded (i.e. (LOC, ORG) when a LOC mention was wrongly predicted as an ORG mention).

We also recorded invalid label transitions for the token-level predictions. 
This was only applicable for XLM-R FLERT, as the other two made span-label predictions instead of token-level.
Table \ref{tab:errors_domain} summarizes the results obtained by the three models across document domains.

\begin{table*}[tb]
    \small
    \centering
    \begin{tabular}{lrrr}
    \toprule
         & Sports & Economy & World Events \\ 
         \midrule
        XLM-R FLERT & &  &  \\ 
        \hspace{0.25cm}Missed & 1 & 14 & 10 \\ 
        \hspace{0.25cm}Spurious & 16 & 49 & 14 \\ 
        \hspace{0.25cm}Boundary Error & 67 & 48 & 20 \\ 
        \hspace{0.25cm}Type Error & 77 & 61 & 16 \\         
        \midrule
        LUKE & &  &  \\ 
        \hspace{0.25cm}Missed & 11 & 39 & 16 \\ 
        \hspace{0.25cm}Spurious & 4 & 28 & 7 \\ 
        \hspace{0.25cm}Boundary Error & 54 & 42 & 22 \\ 
        \hspace{0.25cm}Type Error & 54 & 39 & 22 \\ 
        \midrule
        ASP-T0-3B & &  &  \\  
        \hspace{0.25cm}Missed & 11 & 16 & 8 \\ 
        \hspace{0.25cm}Spurious & 46 & 23 & 7 \\ 
        \hspace{0.25cm}Boundary Error & 124 & 43 & 25 \\ 
        \hspace{0.25cm}Type Error & 61 & 48 & 13 \\ 
        \bottomrule
    \end{tabular}
    \caption{Error types per document domain on CoNLL\#}
    \label{tab:errors_domain}
\end{table*}

\subsection{Recurrent Errors in economy documents}
As introduced in Section \ref{subsection:performance}, all state-of-the-art NER models that we tested performed significantly worse on economy test documents than the others. 
We can also see that these mostly come from data reports and hybrid articles in the economy domain.

Using SeqScore's error counts feature \citep{palen-michel-etal-2021-seqscore}, we counted the mention-level errors of all three models on economy documents (Table \ref{tab:errors}).

\begin{table}[tb]
\small
\centering
\begin{tabular}{rlll}
\toprule
\multicolumn{4}{c}{XLM-R FLERT}\\
\cmidrule(lr){1-4}
Count & Error & Type & Text\\
\midrule
       3 & FP & ORG & Chicago  \\ 
       3 & FN & LOC  & Chicago   \\
       3 & FP & MISC  & Select  \\
       2 & FN & MISC & ACCESS  \\
       2 & FP & ORG & Busang   \\
       2 & FN & LOC & Busang \\
       2 & FN & MISC & Canadian  \\
       2 & FP & MISC  & Choice   \\
       2 & FP & ORG & Busang   \\
       2 & FN & LOC & Busang \\
       2 & FP & ORG & Ministry \\
\midrule
\multicolumn{4}{c}{LUKE}\\
\cmidrule(lr){1-4}
Count & Error & Type & Text\\
 3 & FN & ORG & NYMEX \\
 2 & FN & MISC & ACCESS \\
 2 & FN & MISC & Canadian \\
 2 & FP & LOC & Canadian West Coast \\
 2 & FP & ORG & Durum \\
 2 & FP & MISC & GDR \\
 2 & FN & ORG & Manitoba \\
 2 & FN & ORG & Manitoba Pork \\
\midrule
\multicolumn{4}{c}{ASP-T0-3B}\\
\cmidrule(lr){1-4}
Count & Error & Type & Text\\          
\midrule
 4 & FN & MISC & trans-Atlantic    \\ 
 3 & FN & ORG  & NYMEX     \\
 2 & FP & ORG  & ACCESS \\
 2 & FN & MISC & ACCESS      \\
 2 & FN & MISC & Canadian \\
 2 & FP & LOC & Canadian West Coast       \\
 2 & FP & LOC  & Iowa-S Minn   \\
 2 & FN & MISC  & Iowa-S Minn   \\
 2 & FN & ORG  & Manitoba \\
\bottomrule
\end{tabular}
\caption{Most frequent false positive and false negative errors for each of the three NER models' predictions on economy documents}
\label{tab:errors}
\end{table}

Our manual error analysis revealed that the economy domain documents had a high density of tough mentions, which we classified as follows.

\paragraph{Ambiguous acronyms}
Economy articles were much more packed with acronyms and initialisms that the models struggled with. Acronyms are predominantly a mixture of ORG and MISC mentions.
Examples: \emph{NYMEX}, \emph{ADRs}, \emph{BTPs}, \emph{CEFTA}, \emph{CST}, \emph{CBOT}, \emph{ORE}.

\paragraph{Obscure / unseen mentions}
Much more than in the other domains, economy documents contained mentions which were not only unseen in the training data, but perhaps too rare even for the pre-trained embeddings to help. This is crucially made worse by the fact that they often lacked sufficient context within the sentence or even document. This is especially true amongst economy data reports, which was the document type with worst overall F1 scores. This ranges from commercial product names [\emph{Arabian Light}, MISC], to obscure international companies [\emph{Thai Resource}, ORG], to purely esoteric mentions [\emph{Algoa Day}, MISC (name of a ship)].

\paragraph{Unlikely mentions}
In many cases, tokens that likely have high correlation with one type are present in economy documents with an idiosyncratic type. One clear example of this is the mention \emph{Manitoba}, which co-refers to an organization named \emph{Manitoba Pork} named earlier in the document. All of the models mislabeled \emph{Manitoba} as a LOC in this string:
\emph{Manitoba 's Hog Price Range : 84.00-86.00 per cwt}

\paragraph{Capitalized non-entities}
Table \ref{tab:errors_domain} shows that models committed the most SPURIOUS errors in the economy domain. An example of this type of error occurs in the stock price data reports, in which assets that are not named entities are capitalized, such as in the following string: \emph{Wheat 121 130 121.3 121 Maize (Flint) 113 114 113.7 112 Maize (Dent) 113 114 113.7 112}

According to the section A.4.5.2 of the original MUC guidelines for Named Entities, ``sub-national regions when referenced only by compass-point modifier'' should not be tagged as locations. The CoNLL-03 English corpus stayed true to this directive, and has labeled unnamed regions such as \emph{East Coast} and \emph{West Coast} with the O tag. However, most likely due to their capitalization, all of the models mislabeled them as LOC mentions.

\subsection{Other Recurrent Errors}

\paragraph{Compound mentions}
The most common error type across all models and document types were boundary errors (39.0\%). 
In many cases, the models would get confused by adjacent mentions, and could not parse if they were two distinct mentions, or part of one mention. 
Strings such as \textit{Nazi German}, \textit{Algerian Moslem}, and \textit{UK Department of Transport} are two separate mentions each, but all three models incorrectly treated them as single mentions of two-token length.
Conversely, the mention \emph{1993 World Cup}, all three models incorrectly left off 1993 and tagged the mention as \emph{1993} [\emph{World Cup}]\textsubscript{MISC}.

\paragraph{Irregular capitalization}
Modern state-of-the-art NER systems are still confounded by irregular capitalization cues, at least when trained on the CoNLL-03 English corpus. This is recurrent across all models when dealing with all-caps headlines (all three models labeled \emph{CITY OF HARTFORD} as a LOC, instead of simply HARTFORD), spuriously tagged capitalized non-entities (\emph{Business Policy}), and missed mentions that were in lower case (\emph{world wide web}).

\section{Limitations}
While still a fixture of NER evaluation 20 years later, the CoNLL-03 English corpus remains imperfect as a benchmark. 
Many of the errors that we identified persist within the training and development sets. Though past correction efforts worked to ameliorate these errors \citeplanguageresource{reiss-etal-2020-identifying,Fu_Liu_Zhang_2020}, the overwhelming majority of novel NER models are trained on the original data.

Due to limited resources for analysis and paper length limits, we have only focused on CoNLL-03 English data for this paper.
The CoNLL 2002--3 NER shared tasks included languages beyond English (Dutch, German, and Spanish).
Our early investigations showed that the other languages also show systematic, impactful annotation errors.
In particular, as previously noted by \citet{Agerri2016-fy}, the CoNLL-02 Spanish test set has many mentions that include bordering quotation marks, which goes against common NER conventions. 

It is just as important for NER models in other languages to be tested on data that is as consistent and clean as possible, so that we can learn more about the lingering NER errors in languages beyond English.
For example, our analysis of the CoNLL-02 Spanish data found that state-of-the-art models still struggle with parsing the preposition ``de'' within mentions.

There are of course many NER datasets beyond the CoNLL 2002--3 shared tasks, and those recently developed for less-resourced languages are of particular interest to us.
We hope to collaborate with speakers of those languages to extend our study far beyond English.

Finally, a lingering issue in NLP evaluation is to what extend the high results obtained by models may be caused by data contamination. 
It is likely that the models saw the CoNLL-03 English data during pretraining, which may explain the the high results obtained on this task.
Exactly what this means for our evaluation is a matter of debate, and the impact that data contamination may be having on evaluation is currently an ongoing line of research \cite{sainz-etal-2023-nlp}.

\section{Conclusion}

In this paper, we conducted a full-scale error analysis of state-of-the-art named entity recognition taggers on the CoNLL-03 English dataset. Our document type annotations revealed clear trends, including the relatively poor performance that state-of-the-art models achieve on the 67 economy documents in the test set.
We evaluated and adjudicated the various corrected English CoNLL test sets to create CoNLL\#, a version of the CoNLL-03 English test set with significantly less annotation-error noise.
Finally, we tested three state-of-the-art NER models on this corrected test set, and combed through their errors to get a full sense of what they continue to struggle with, and detail where future models can gain those last F1 points.

\section{Ethics and Broader Impact}

Benchmark datasets play a key role in NLP research.
Improvements in benchmark results are generally accepted as overall progress on a given task. 
However, this can also lead to benchmark chasing, which reduces the difficulty of a given task to a matter of gaining tenths of a point on a leaderboard, without truly gaining any insight or making true advancements on the task \citep{raji2021}. 

In addition, no benchmark can ever fully capture the complexities of the linguistic phenomenon in question \citep{paullada2021data}.
Equating NER progress simply to gains in F1 score on a given benchmark is a reductionist approach.
As popular as CoNLL-03 English may be, it suffers from obvious limitations: the genre of the documents are exclusively newswire texts covering a limited set of topics (sports, economy, world events) from a short span of time (1996) and only certain language varieties are represented.

We hope the broader impact of this work will be that progress on the CoNLL-03 English dataset can be better measured due to a lower-noise version of the test set and that others will be able to adapt our methodology to other datasets and languages.

\section{Bibliographical 
References}\label{reference}

\bibliographystyle{lrec-coling2024-natbib}
\bibliography{main}

\section{Language Resource References}
\label{lr:ref}

\bibliographystylelanguageresource{lrec-coling2024-natbib}
\bibliographylanguageresource{main_languageresource}

\appendix
\section{Data statement}
\label{sec:data_statement}
We document the information concerning CoNLL\# following the data statement format proposed by \citet{10.1162/tacl_a_00041}.\\ 
\textbf{Data set name:} CoNLL\# \\
\textbf{Data set developers:} Andrew Rueda, Elena Álvarez Mellado, and Constantine Lignos \\
\textbf{Dataset license:} Our modifications to the original data are distributed under \href{https://creativecommons.org/licenses/by/4.0/}{CC BY 4.0}. The original license terms apply to the original data. 
\\ 
\textbf{Link to  dataset:} \url{https://github.com/bltlab/conll-sharp}

\subsection{Curation rationale}
CoNLL\# is a reannotation and adjudication of the English section of the CoNLL-03 test set. 

\subsection{Language variety}
The language of this corpus is English (ISO 639-3 \texttt{eng}), of the variety used in international newswire. 

\subsection{Speaker demographic}
No detailed information was collected regarding the demographics of the authors of the original texts from CoNLL-03.
However, we can infer that the authors of the text were English-speaking journalists aged between 20-65.

\subsection{Adjudicator demographic}
The annotator and adjudicator of CoNLL\# was a 20-30 year-old male graduate student from the USA, trained in linguistics and computational linguistics, whose native language is English.

\subsection{Speech situation}
\label{sec:speech}
The English section of the CoNLL-03 dataset is taken from the Reuters Corpus, which consists of a collection of English journalistic texts written between 1996 and 1997. 
For a full description of the CoNLL-03 dataset see \citetlanguageresource{tjong-kim-sang-de-meulder-2003-introduction}.

\subsection{Text characteristics}
The texts from CoNLL-03 English come from the Reuters Corpus. 
This means that the texts in CoNLL\# are from the newswire domain.
Consequently, we can assume that all the texts in CoNLL\# are carefully, well-edited texts that follow the rules of ``standard'' English.

The articles in the test set belong to the sports domain, economy or world events (see section \ref{subsection:document_domains}).
In terms of format, the documents in CoNLL\# are text articles, data reports (tabulated data) or a mix between the two (see section \ref{subsection:document_formats}).

\subsection{Recording quality}
N/A
\subsection{Other}
N/A

\end{document}